\documentclass{llncs}[11pt]

\usepackage{dcolumn,listings}
\usepackage{endnotes,mathtools,amsmath,amssymb}
\usepackage[ruled,vlined]{algorithm2e}

\newcommand{\ei}{EI}
\newcommand{\eis}{EIs}

\newcommand{\naturals}{\mbox{\ensuremath{\mathbb{N}}}}
\newcommand{\reals}{\mbox{\ensuremath{\mathbb{R}}}}

\newcommand{\I}{\ensuremath{\mathcal{I}}}

\begin{document}


\title{Social Contract Negotiation\\  \vspace{5mm}\small Open Standardization of Social Inclusion and Equity.}

\author{Andr\'es Garc\'ia-Camino \\ dragc@duck.com}\institute{Independent Researcher, Barcelona, Spain}

\maketitle


\begin{abstract}
Social choice is the theory about collective decision towards social welfare starting from individual opinions, preferences, interests or welfare. The field of Computational Social Welfare is somewhat recent and it is gaining impact in the Artificial Intelligence community. Classical literature makes the assumption of single-peaked preferences, \emph{i.e.} there exist a linear order in the preferences and there is a global maximum in this order. Recently, some theoretical results were published about Two-stage Approval Voting Systems (TAVs), Multi-winner Selection Rules (MWSR) and Incomplete (IPs) and Circular Preferences (CPs) that I claim leads to research about Preferences Graphs and Preferences Multi-dimensional Functions in Polynomial time. 

The purpose of this paper is three-fold: Firstly, I want to introduce Social Choice Optimisation as a generalisation of TAVs where there is a maximization stage and a minimization stage implementing thus a Minimax, a well-known Artificial Intelligence decision-making rule to minimize hindering towards a (Social) Goal. Secondly, I want to introduce, following my Open Standardization and Open Integration Theory (in refinement process) put in practice in my dissertation, the Open Standardization of Social Inclusion, as a global social goal of Social Choice Optimization.

\end{abstract}

\section{Motivation}

Social choice is the theory about collective decision towards social welfare starting from individual opinions, preferences, interests or welfare. The field of Computational Social Welfare is somewhat recent and it is gaining impact in the Artificial Intelligence community. Classical literature makes the assumption of single-peaked preferences, \emph{i.e.} there exist a linear order in the preferences and there is a global maximum in this order. Recently some theoretical results were published about Two-stage Approval Voting Systems (TAVs), Multi-winner Selection Rules (MWSR) and Incomplete (IPs) and Circular Preferences (CPs) that I claim leads to research about Preferences Graphs and Preferences Multi-dimensional Functions in Polynomial time. 

The purpose of this paper is three-fold: Firstly, I want to introduce Social Choice Optimisation as a generalisation of TAVs where there is a maximization stage and a mininimization stage implementing thus a Minimax, a well-known Artificial Intelligence decision-making rule to minimize hindering towards a (Social) Goal. Secondly, I want to introduce, following my Open Standardization and Open Integration Theory (in refinement process) put in practice in my dissertation \cite{garcia2010normative}, the Open Standardization of Social Inclusion, as a global social goal of Social Choice Optimization.

As for the Open Standardization of Social Inclusion, I will start finding an open consensus between Adaptive Coordinate descent and Coherence Theory, extending the propositional evaluation of the latter for the multi-variate functional case, adjusting thus both domains. Then I will continue towards an open integration of both approaches introducing the Coherent Social Inclusion Problem. Finally, I will provide an algorithm for the One-stage Approval Voting (OAV) and two for the Two-stage Approval Voting (TAV), namely, Preference Number Maximization (PNM) and Preference Management (PM). PNM is useful for tackling the overall known problem, but is NP-hard in the uncertain case as it is a generalisation of the Travelling Salesman Problem for continuous Halmiltonian Paths. Let me argue this statement. Adaptive Coordinate Descent is a generalisation of Gradient Descent which in turn it is a generalisation of Hill Climbing.

That is my point, imagine that you and some other strangers got kidnapped and left unconscious in the middle of a mountain. Let also assume, that you want to reach a peak to see where are you in order to continue with the escape plan. (Rolling downwards is obviously disregarded). In that case, you might want to follow the most gradual and feasible path minimising the distance walked too. Having a bit of knowledge of the mentioned mountain or having a map would equal to finding the gradual shortest path. However, without any knowledge of the mountain, only with an intuition about the goal, you and your luckily friendly partners should arrive to a consensus about the paths to follow at any bifurcation. One heuristic would be to analyse a subset of all the possible paths, choose the k-gradual and feasible paths and vote the final solution. That would be Coherent Social Inclusion.

The structure of this paper is as follow: Section \ref{sec:social-choice} introduces the field and latest research results on Social Choice. A generalization of Coherence Theory is introduced in section \ref{sec:coherence-theory}. The first approach presented in this paper is Two-stage Approval Voting Optimization and it is enunciated in section \ref{sec:TAV-optimization}. Secondly, the minimization of Discriminating Preferences is introduced in section \ref{sec:minimization}. Thirdly, the Coherent Social Inclusion $CSI$ Problem is defined in section \ref{sec:coherent-inclusion}. The overall process for one and two stages without uncertainty, namely, is presented in section \ref{sec:no-uncertainty}.  Finally, in  section \ref{sec:policy-making} $CSI$ is adapted for a Policy-making scenario.

\section{Context}

\subsection{Social Choice}\label{sec:social-choice}
Social choice is the theory about collective decision towards social welfare starting from individual opinions, preferences, interests or welfare. The field of Computational Social Welfare is somewhat recent and it is gaining impact in the Artificial Intelligence Community. Classical literature makes the assumption of single-peaked preferences, \emph{i.e.} there exist a linear order in the preferences and there is a global maximum in this order. Recently some theoretical and polynomial time results were published about Two-stage Approval Voting Systems (TAVs), Multi-winner Selection Rules (MWSR) \cite{aaai18:polynomial-MW-elections} and Incomplete (IPs) \cite{incomplete-prefs2020} and Circular Preferences (CPs) \cite{circular-prefs2016} that I claim leads to research about Preferences Graphs and Preferences Multi-dimensional Functions in Polynomial time.

\subsubsection{Two-stage Approval Voting}
Approval Voting  (AV) is a single-winner electoral system where each voter may select $k$-candidates. The winner is the most-approved candidate. There are extensions for the selection of $l$-winners using Multi-winner Selection Rules (MWSR) in Proportional Approval Voting:

\begin{definition}\label{def:MWSR} 
`Let us now give and analyse our IP formulation for PAV.
This formulation has one binary variable $y_c$ for each candidate
$c \in C$, indicating whether candidate $c$ is part of the
committee. Constraint (2) requires that the committee contains
exactly $k$ candidates. The binary variables $x_{i,l}$ indicate
whether voter $i \in N$ approves of at least $l$ candidates in the
committee; this interpretation is implemented by the constraints
(3)'
\begin{eqnarray}
    maximise &\sum_{i\in N}\sum_{l\in[k]}\alpha_l\cdot x_{i,l}& \\
    subject\ to &\sum_{c\in C} y_c = k& \\
    &\sum_{l\in[k]}x_{i.l}\leq\sum_{i\ approves\ c} y_c &for\ i\in N\\
    &x_{i,l} \in \{0,1\} &for\ i\in N, l\in[k]\\
    &y_c \in \{0,1\} &for\ c \in C
\end{eqnarray}
Extracted from  \cite{aaai18:polynomial-MW-elections}.
\end{definition}

Two-stage Approval Voting (TAV) is a refinement of AV, where the global Selection Rule is divided into two selection rules that narrows the group of candidates passing to the next stage.

\subsection{Coherence Theory}\label{sec:coherence-theory}

Thagard proposed a decision-making Theory of Coherence about a graph of atomic propositional preferences with a computational coherence function that it is modified when newer preferences are added \cite{coherence02}.

\section{Preliminary Proposals}

The author proposes to specify Coherence Theory with a multi-dimensional Coherence function is each edge of the Graph.

\subsection{Optimizing Two-stage Approval Voting}\label{sec:TAV-optimization}

Following the classical approach to TAVs, the MWSR selects the k-winners with the most votes in both stages. On the contrary, the Author proposes first a maximisation (most positive votes) stage and then a minimisation (less negative  votes) stage. Thus, applying the Minimax Condorcet Method for the optimisation of Social Choice.

\subsection{Discriminating Preferences Minimization}\label{sec:minimization}

Taking from granted Social Inclusion as the global goal of Social Choice Optimization, the author defines
Social Discrimination as the inverse of a multi-dimensional Utility function representing the Social Inclusion of an agent in a Society. 

As for minimization, Adaptive Coordinate Descent\cite{ACD2011} is a Gradient Descent generalisation for optimization of non-derivable functions, as the Social Discrimination functions.

\subsubsection{Declarative Electronic Institutions may Discriminate}

Any type of constraint in the Social Universe (either physical, social, normative, moral, ethic, etc.) are implicitly generalized as the Social Discrimination function.

For example, the author envisages several Social Discrimination axis, namely, the agreed (physical, social, normative, moral, ethic$\ldots$) categories leading exclusion and discrimination risk, \emph{e.g} age (range), gender, ethnicity,\ldots,negative permissions, prohibitions, obligations, duties, power and so on.  

In the following applied example I will use Declarative Mechanism Design \cite{agc:aicomm20-2}:

\[ DEI_t:  Events_t \times Rules_t \times InstEvents_{t-1} \rightarrow InstEvents_{t+1} \]

A Declarative Electronic Institution calculates the aggregation and removal of agreed events ($InstEvents$) for a set of previous $InstEvents$ (possibly empty), a set of Rules and a set of Events.

Assuming a declaration order and using Social Discrimination definition it would be defined as follows:

\[ DEI_t = \overrightarrow{SD}: Events_t \times InstEvents_{t-1} \rightarrow 2^\mathbb{B} \]

A Declarative Electronic Institution assigns a Boolean on the acceptance on the aggregation of some Events at time $t$ depending on the previous ones.

Generalizing,the Institutional axis might be:

\[ Axis(DEI_t) = \overrightarrow{SD}: Event \rightarrow \mathbb{B} \]

Nevertheless, events occur concurrently and are processed in tandem, so:

\[ Axis(DEI_t) = \overrightarrow{SD}: Events \rightarrow 2^\mathbb{B} \]

A Declarative Electronic Institutions determines the acceptance of concurrent Events. 

Thus, s Social Discrimination function that all the Declarative Electronic institutions might implement would be:

\[ \overrightarrow{SD}: Society\times Agents\times Events\times Time \rightarrow [0,1]\]

That is, it studies the discrimination degree of agent events over time with regard of a Society.

\begin{definition}\label{def:SD}
A Social Discrimination function $\overrightarrow{SD}$ determines the degree of non acceptance of a set of preferences of a subset $A_p$ of agents with respect of a Society over time:

\[ \overrightarrow{SD}_n: Society\times Agents\times \overrightarrow{SD_0} \times \ldots \times \overrightarrow{SD}_n \times Time \rightarrow [0,1]\]

where $n$ are the (possibly infinite) dimensions in $\overbrace{DS}$, the Discrimination Space and: \[\overrightarrow{SD_1}:Society\times Agents \times P\times Time \rightarrow[0,1]\]

where $\overrightarrow{SD_1}$ calculates the agreed discrimination degree of a set of Preferences $P$ and;

\[\overrightarrow{SD_0}:Society\times Agents \times T\times Time \rightarrow[0,1]\]

where $\overrightarrow{SD_0}$ calculates the agreed discrimination degree of a set of Traits $T$.
\end{definition}

\begin{definition}\label{def:SD}
A Knowledge Map $\overbrace{KM}$ is a (partial and globally agreed) implementation of $\overrightarrow{SD}$ in $n$ dimensions of the Social Discrimination space $\overbrace{DS}$ such that:
\[\overbrace{KM}:\overrightarrow{SD}\rightarrow \mathbb{U} \times \mathbb{D}\]
where $\mathbb{U}$ is the degree of Uncertainty and $\mathbb{D}$ is the degree of Social Discrimination.
\end{definition}

For instance, Gravity Law on Earth has $\mathbb{U}=0$ and $\mathbb{D}=0$ since it is world-wide agreed that limits all (Human) Societies equally. Similarly, Absolute Majority rule has $\mathbb{U}<<1$ although $\mathbb{D}>>0$ since it is the world-wide accepted Decision Rule even if it discriminates other preferences and candidates in the decision process.

Let me then introduce the Social Inclusion Problem and a characterization of its possible solutions.

\section{Coherent Social Inclusion}\label{sec:coherent-inclusion}

\subsection{Coherent Social Inclusion Problem}

\begin{definition}\label{def:SIP}

For a Social Universe $SU=\langle A, S, SIP\rangle $ where there exist $A$, the set of all agents $a_i$ in $SU$,  a set of Society functions $\overrightarrow{S}$ such that $\overrightarrow{S}:A \rightarrow 2^A$,  and a set of Social Inclusion Problems $SIP^n$ in $n$ dimensions, such that the i-th problem:

\[SIP_i=\langle \underbrace{SPwr}, \overrightarrow{SU}, T, SDP,  P, \overrightarrow{SP}, \overrightarrow{SD} \rangle\]

where:

A Social Power order $\underbrace{SPwr}$ between Socities due to the Social Utility $\overrightarrow{SU}$ calculated as the sum of weighted average Utilities (of presumed traits) of all the agents in the Society,
$T$ is the set of all presumed traits of the set of agents $A$. $SDP$ is the set of Social Discrimination Profile functions in $n$ dimensions over $T$ for all the agents in $A$. $P$ is the set of all preferences $p$ in $SIP$. $\overrightarrow{SP}$ is the Selected Preferences function such that $\overrightarrow{SP}:P\rightarrow2^P$; namely the subset of preferences taken into account for preference aggregation. Finally, $\overrightarrow{SD}^n$ is the Social Discrimination function over $P$ in $n$ dimensions,as per definition \ref{def:SD}.

The Social Discrimination Profile functions $SDP$ are discrimination estimations agreed upon previously over the Social Roles of a Society. Namely, they are a set of vectors $\overrightarrow{d}$ in the Social Discrimination Space.
\end{definition}

\begin{definition}\label{def:agent}
As for agency, an agent $a\in A$:
\[A_i=\langle  T_i, SDP_i,\overrightarrow{K}_i,\overrightarrow{P_i}, PD_i \rangle\]
where $t_i$ is the trait $t$  of agent $i$; $T_i$ is the set of all the traits of agent $i$, where $\overrightarrow{U}_i\in T_i$ is the Utility trait for agent $i$, 
$SDP_i^n$ is the Social Discrimination Profile function of agent $i$ in $n$ dimensions, such that $SDP_i^n:T_i\rightarrow \reals^n$. Moreover, $K_i$ is the (Possible and Coherent) Preference Knowledge of agent $i$ such that $K_i:P\times \overrightarrow{SD} \rightarrow 2^P\times 2^{\overrightarrow{PD}_i^n}$ where each $PD_i^n$ is a Preference Discrimination function of agent $i$ in $n$ dimensions such that $\overrightarrow{PD}_i^n:\overrightarrow{P_i} \times P \rightarrow [0,1]$; $p_i$ is the preference $p$ of agent $i$. Subsequently, $\overrightarrow{P_i}$ is the Preference function of agent $i$ such that $\overrightarrow{P_i}:\overrightarrow{K}_i \times P \rightarrow2^{P}$. 
\end{definition}

\subsection{Coherent Social Inclusion Solutions}

\begin{definition}
$\underbrace{S}$ is a solution for a Social Universe $SU$ that finds a Coherent Inclusion Problem $i$ that maximises Social Power $\underbrace{SPwr}$ and minimizes Social Discrimination $\overrightarrow{SD}$.
\end{definition}

\begin{definition}
Coherent Policy-Making $\overrightarrow{CPM}$ is a function:
\[ \overrightarrow{CPM}: SU_t \times SU_{t-1} \rightarrow \underbrace{S_t} \]
where $\underbrace{S_t}$ is a solution of Coherent Social Inclusion at time $t$.
\end{definition}

\begin{definition}
Optimal Policy-Making $\overrightarrow{OPM}$ is a function:
\[ \overrightarrow{OPM}: SU_t \times SU_{t-1} \rightarrow \underbrace{S_t} \]
where $\underbrace{S_t}$ is the best solution of Coherent Social Inclusion at time $t$.
\end{definition}

\subsection{Discrimination}

The notion of social discrimination has been defined thoroughly from a qualitative perspective. In this paper, the author wants to create a quantitative definition of social discrimination as follows:

\begin{definition}
Social Discrimination are the limitations of a (sub)society imposed by another society. It may also use Graded Ostracism, namely, forcing demote to a sub-society in the Social Power order.
\end{definition}

In definition \ref{def:SIP} Social Power $\underbrace{SPwr}$ orders sub-societies for their Social Utility $\overrightarrow{SU}$ introduced as the sum of the weighted average $Utility_{T_i}$ \ of (pressumed traits $T_i$) for all agents $i$ in the sub-society. This weighted average may vary from sub-society and generalises Social Prejudices, like racial, sexual orientation and so on.

\begin{definition}
When the number $s$ of sub-societies tends to infinity, Social Power function is defined as:
\[\underbrace{SPwr^{s\rightarrow\infty}}=\overrightarrow{SPwr}:\overrightarrow{S}\rightarrow [0,+\infty]\]
for a society $\overrightarrow{S}$ is the super-society of Social Universe $SU$.
\end{definition}

\begin{definition}
When the number $s$ of sub-societies tends to infinity, Social Utility is defined as:
\[\overrightarrow{SU}^{s\rightarrow\infty}:\overrightarrow{S}(S) \rightarrow +\infty\]
where $S$ is the maximal society of Social Universe $SU$, a super-society.
\end{definition}

\begin{definition}
When the number $s$ of sub-societies tends to infinity, :
\[\underbrace{SPwr^{s\rightarrow\infty}}=\overrightarrow{SU}^{s\rightarrow\infty}(S)=+\infty\]
for $S$ the maximal society of Social Universe $SU$, the super-society.
\end{definition}

\begin{definition}
When  $s=0$, Social Power function is defined as:
\[\underbrace{SPwr^{s=0}}=\overrightarrow{SPwr}(\overrightarrow{s})\rightarrow +\infty\]
where $\overrightarrow{S}=\overrightarrow{s}$ is a the minimal society of Social Universe $SU$ formed only by a agent.
\end{definition}

\begin{definition}
When  $s=0$, Social Utility is defined as:
\[\overrightarrow{SU}^{s=0}: \overrightarrow{S}(\overrightarrow{s}) \rightarrow +\infty\]
where $\overrightarrow{S}=\overrightarrow{s}$ is a the minimal society of Social Universe $SU$ formed only by a agent.
\end{definition}

\begin{definition}
When  $s=0$ :

\[\underbrace{SPwr^{s=0}}=\overrightarrow{SU}^{s=0}(\overrightarrow{S}(\overrightarrow{s}))=+\infty\]

where $\overrightarrow{S}=\overrightarrow{s}$ is a the minimal society of Social Universe $SU$ formed only by a agent.
\end{definition}

\subsection{Capitalism}

\begin{definition}
When $s=0$, Utility for all agent $i$ in Social Universe tends to infinity, $ \overrightarrow{U_i}\rightarrow\infty$ :

\[\underbrace{SPwr^{s=0}}=\overrightarrow{SU}^{s\rightarrow0} = U_i\] 

and 

\[\underbrace{SPwr^{s=0}}\rightarrow\infty\]

where $\overrightarrow{s}$ is a minimal society of Social Universe $SU$ formed only by a agent.
\end{definition}

\begin{definition}
When $s\rightarrow\infty$, Utility for all agent $i$ in Social Universe tends to infinity, $ \overrightarrow{U_i}\rightarrow \infty$ :

\[\underbrace{SPwr^{s\rightarrow\infty}}=\overrightarrow{SU}^{s\rightarrow\infty} = U_i\] 

and 

\[\underbrace{SPwr^{s\rightarrow\infty}}\rightarrow \infty\]

where $\overrightarrow{S}$ a the maximal society of Social Universe $SU$, a super-society.
\end{definition}

\begin{definition}
When $s\rightarrow\infty$, Utility for all agent $i$ in Social Universe tends to infinity, $ \overrightarrow{U_i}\rightarrow 0$ :

\[\underbrace{SPwr^{s\rightarrow\infty}}=\overrightarrow{SU}^{s\rightarrow\infty} = U_i\] 

and 

\[\underbrace{SPwr^{s\rightarrow\infty}}\rightarrow 0\]

where $\overrightarrow{S}$ a the maximal society of Social Universe $SU$, a super-society.
\end{definition}

\begin{definition}
When $ \overrightarrow{U_i}\rightarrow \infty$ :

\[\underbrace{SPwr}=\overrightarrow{SU} \rightarrow \infty\]

\end{definition}

\begin{definition}
When $ \overrightarrow{U_i}\rightarrow 0$ :

\[\underbrace{SPwr}=\overrightarrow{SU} \rightarrow 0\]

\end{definition}

That shows some Properties of Capitalism, under Social Power and Social Discrimination, and leads to try to find a compromise of Social Discrimination versus Societal Autonomy in a Policy-making scenario.

This may be be achieved by decreasing the weight of Utility on the contribution to Social Utility, may be remunerating each vote that contributes to the Society, as Participatory Utility $\overrightarrow{PU}_i$.
Concretely, by using Linear Programming with these constraints and the following methods:

\section{CSI Algorithms without Uncertainty}\label{sec:no-uncertainty}

\subsection{One-stage Approval Voting}

 $\overrightarrow{CSI^1}$ is the One-stage Coherent Social Inclusion function for a Society such that maximizes by voting the set of agents such that there exist a Inclusive Multi-winner Selection Rule function:
    \[IMWSR:A\times\overrightarrow{S}\times SIP_i^n \rightarrow CSIS \] where $CSIS$ is the Coherent Social Inclusion Society such that maximises by voting the number of agents belonging to $S$ such that $k\in\naturals$: \[\forall a_i\in\overrightarrow{S}(A),\forall T_i: \overrightarrow{SDP}(T_i)=argmin_{k}\ \overrightarrow{SD}(P)\].

That is, we would like to maximize the number of agents with the minimum traits being discriminated for.
For the calculation of this One-stage Coherent Social inclusion function, we firstly apply the Adaptive Coordinate Descent algorithm to find a set of less socially discriminatory all the preferences P, and then we apply the one-stage of the Approval Voting system where we maximize the number of winners with no trait being discriminated for.

\subsection{Two-stage Approval Voting}

\subsubsection{Preference Number Maximization}

$\overrightarrow{CSI^2}$ is the Two stage Coherent Social Inclusion function for a Society such that maximizes by voting the set of agents such that there exist a Two-stage Inclusive Multi-winner Selection Rule function:
     \[TS-IMWSR:A\times\overrightarrow{S}\times SIP_i^n \rightarrow CSIS \] where $CSIS$ is the Coherent Social Inclusion Society such that maximises by voting  the number of agents belonging to $S$ such  that $j,k,l\in\naturals,l>j>k$:
    
    \begin{description}
    \item[Stage one] \[Stage^1=MWSR(l,P)\]
    \item[Stage two] \ \\\ \\1. 
        $$\forall a_i\in\overrightarrow{S}(A),\forall T_i: m = argmin_j\  \overrightarrow{SDP_i}(\overrightarrow{SD}(Stage^1)) $$\\
        3. $$Stage^2= MWSR(k,m)$$
    \end{description}

That is, we would like to maximize the number of agents with the minimum traits being discriminated for.
For the calculation of this Coherent Social inclusion function, we firstly use a Multiple Winner Selection Rule, $MWSR$, on all preferences $P$, as per definition \ref{def:MWSR}. Then we apply the Adaptive Coordinate Descent algorithm on the composition of the Social Discrimination function and Social Discrimination profile of the result of previous stage. Finally,  $MWSR$, namely the Multi-Winner Selection Rules of the second stage is used to find a set of less socially discriminatory winning preferences. 

If the number of agents $i$ were very large, this algorithm would perform poorly. That is the main reason for the next and final algorithm, suitable for Policy-making.

\section{Social Choice Optimization in Policy-making}\label{sec:policy-making}

In this Policy-making scenario, the author acknowledge some degree of Uncertainty in the Social Discrimination function $\overrightarrow{SD}$ without entering in the Agency Uncertainty scenario where the traits of the agents are unknown.

In this section, although two possibilities are considered, namely, Preference Aggregation (Addition) and Preference Derogation (Removal), later on we will see that both cases are forward-chaining, renaming thus the algorithm to Preference Management.

\subsection{Preference Aggregation}\label{sec:PM}

Assuming that a Society already exists and starts with a Coherent Inclusion Problem and minimum set of norms including a Coherent Social Inclusion function, one could assume that if the norms are already Socially Coherent, namely, they minimize the discrimination of the Society. Then the objective is to apply the Coherent Social Inclusion function to add new preferred norms that continues coherently including the whole society.

$\overrightarrow{PACSI^2}$ is the Two-stage Preference Aggregation Coherent Social Inclusion function for a Society such that maximizes by voting the set of agents such that there exist a Less-discriminatory Multi-winner Selection Rule function:
     \[LDM-WSR:A\times\overrightarrow{S}\times SIP_i^n \rightarrow CSIS \] where $CSIS$ is the Coherent Social Inclusion Society such that maximises the number of agents belonging to $S$ such that $j,k,l\in\naturals,l>j>k$:

    \begin{description}
    \item[Stage one] \[Stage^1=MWSR(l,\overrightarrow{SP(}P))\]
    \item[Stage two] \ \\\ \\1. $$  m =argmin_j\ \overrightarrow{SD}(Stage^1)$$\\
        2. $$ p_i = ShortesthPath(\overrightarrow{SD},Stage^1,m)$$\\
        3. $$ Stage^2_i=MWSR(k,p_i)$$
    \end{description}

That is, we would like to maximize by voting the number of agents with the minimum traits being discriminated for.
For the calculation of this Coherent Social inclusion function, we firstly use a Multiple Winner Selection Rule, $MWSR$, on a subset of all preferences $\overrightarrow{SP}(P)$, to start with the aggregation process. As it should return a coherent subset of $l$ preferences, even a random subset might be used to start the initialization process.  Then we apply the Adaptive Coordinate Descent algorithm on the Social Discrimination function $\overrightarrow{SD}$ of the result of previous stage; and then the second-stage $MWSR$ to find a set (of size $k$) of less socially discriminatory winning preferences. 

For this, once found the less discriminatory goal(s) as the single-peak preference(s) we then may find the shortest path(s) in $\overrightarrow{SD}$, a graph with a multi-dimensional discrimination function in each edge. In order to do this, each cost in the graph can be calculated by the application of the discrimination function from vertex to vertex. With the shortest path(s), \emph{i.e.} the optimum (set of) single-peaked preferences, we may now apply the second voting stage (possibly categorised by each path of the set) to find the k-Coherent Optimum Preferences to aggregate.

\subsection{Preference Derogation}

When dealing with Uncertainty and Local Minima, it might be necessary to remove some previously agreed preferences and continue with the Aggregation process. It would be extremely useful in a Policy-making scenario.

In practice, the cost of adding or removing a Preference are different, and may involve more than one Preference depending on the direction. In this case, as a matter of fact, paths in the Social Discrimination function cannot be transited backwards, as generally, there are some actions that does not have inverse, or they are irreversible.

Then, as for the metaphor, the map of the mountain is moreover a directed Graph, like in Vehicle one-direction maps. Thus, we need to redefine our Preference graph to be directed and with possibly different Social Discrimination functions associated with each direction between two vertices.

Thus, the PA algorithm could be renamed as Preference Management $PM$ algorithm in this case, since we can only advance through the graph towards the Global Social Inclusion goal, possibly not optimally if derogation is necessary and mainly due to non-optimal collective decisions in Stage 2 of $\overrightarrow{PACSI}^2$.

\subsection{Compacting the Social Discrimination function}

After a long process of Preference Aggregation and Derogation, the history of the travelled path may contain circles, namely returning to a previous edge in the graph. Thus, for compacting the Preference History one may remove cycles when reaching a previous transited point to obtain a Preference Set. However, it is not desirable to decrease Certainty in a a priori Uncertain function as Social Discrimination $\overrightarrow{SD}$ in order to optimally and automatically learn from errors with the method proposed.

\subsection{Preferences over Social Discrimination functions}

The Coherent Social Inclusion Problem $CSIP$ is recursive by nature: agents need to Coherently Agree in a Social Discrimination function first. So the General Problem is: Could we arrive to agreements on our Preferences over the Agreed Social Discrimination function? A feasible solution to kick-start the method is to apply the method over Preferences over Social Discrimination functions based on current Social Discrimination functions: the definition of our States, Constitutions, and so on.

In a Society initialization case, it may start empty and the selected preferences to vote be chosen randomly, bootstrapping thus in a Brainstorming fashion.

\subsection{Uncertainty in the Social Discrimination Space}

In the Coherent Social Inclusion Problem presented at the beginning of section \ref{sec:coherent-inclusion} there were additional definitions such as $\overrightarrow{PD}$. It was left on purpose to make noticeable the case where the Social Discrimination Function, the map $\overrightarrow{SD}$ of the mountain $\overbrace{DS}$ (the Social Discrimination Space), is not completely known in advance. Thus, it would require to coherently and iteratively aggregating Preference Discrimination Functions of agents, namely, some $\overrightarrow{PD}_i$ for all agent $i$. 

One tentative solution involving Regulated Deep Learning would be using $PM$ in a Regulated Learning phase (Max-phase) where some agreements are made when learning, and employ concurrently $PM$ in a Regulated Decision phase (Min-phase), as the one presented in section \ref{sec:PM} for establishing the Inclusive and Coherent Preference Agreements under Partial Uncertainty.

\section{Justification Example}

To further exemplify, the proposal of this paper, let assume the problem of creating traffic signals in a whole country. We will start assuming two sub-societies in the country, the pedestrian-only and the car-drivers ones:

\[S=\{car_1,\ldots,car_c\}\ and\ s=\{pedestrian_1,\ldots,pedestrian_p\} \]

For brevity, we will also assume that there are four proposals to be voted: $none$,$cross-walks$,$traffic-lights$ and $mixed$ where, respectively correspond to the cases with no signal, create only cross-walks, create only traffic lights and mixed pre-solved approaches are:

\[SIP=\{none,cross-walks,traffic-lights,mixed\}\]

After recursive application of the proposed method, we will also assume we arrive to the following considerations: 
\begin{enumerate}
    \item for\ $SIP=none$: \[SPwr(S) >> SPwr(s),\ SD(cars)<SD(pedestrians) \]
    \item for\ $SIP=cross-walks$: \[ SPwr(S) > SPwr(s),\ SD(cars)<SD(pedestrians) \]
\item for\ $SIP=traffic-light$: \[ SPwr(S) < SPwr(s),\ SD(cars)>SD(pedestrians) \]
\item for $SIP=mixed$: \[ SPwr(S)=SPwr(s),\ SD(cars)=SD(pedestrians)\]
\end{enumerate}

For a toy example like this, it may seem obvious that the better approach is $mixed$ in terms of equality of Social Power and Social Discrimination. However there might be societies where $c>>p$ (cities) or $c<<p$ (small villages) such that the Absolute Majority Rule would not work optimally regarding Social Inclusion, since in those cases $none$ and $traffic-lights$ would win a priori win due to the population unbalance. Especially for those cases, the Less-discriminatory Majority Winner Selection Rule $LDM-WSR$ will optimize policy-making.

\section{Conclusions}\label{conclusions}

In this paper is I have introduced Social Choice Optimisation as a generalisation of Two-stage Approval Voting (TAV) where there is a maximization stage and a  minimization  stage  implementing  thus  a  Minimax, a well-known  Artificial  Intelligence decision-making rule to minimize hindering towards a (Social) Goal. 
Secondly, I have presented, following my Open Standardization and Open Integration Methodology (in refinement process) I put in practice in my dissertation [1], the Open Standardization of Social Inclusion, as a global social goal of Social Choice Optimization. Any type of constraint in the Social Universe (either physical, normative, moral, ethic, etc) are implicitly generalized in the Social Discrimination function.

As for the Open Standardization of Social Inclusion, I started with a open consensus between Adaptive Coordinate descent and propositional Coherence Theory, extending the latter for the functional case. Then I continued towards an open integration of both approaches introducing the Coherent Social Inclusion Problem. Finally, I provided an algorithm for the One-stage Approval Voting (OAV) and two for the Two-stage Approval Voting (TAV), namely, Preference Number Maximization (PNM) and Preference Aggregation (PA). PNM is useful for tackling the overall known problem, but is NP-hard in the uncertain case as it is a generalisation of the Travelling Salesman Problem for continuous Halmiltonian Paths. Let me argue this statement. Adaptive Coordinate Descent is a generalisation of Gradient Descent which in turn it is a generalisation of Hill Climbing. That is my point, imagine that you and some other strangers got kidnapped and left unconscious in the middle of a mountain. Let also assume, that you want to reach a peak to see where are you in order to continue with the escape plan. (Rolling downwards is obviously disregarded). In that case, you might want to follow the most gradual and feasible path minimising the distance walked too. Having a full of knowledge of the mentioned mountain or having a map would equal to finding the gradual shortest path. However, without any knowledge of the mountain, only with an intuition about the goal, you and your luckily friendly partners should arrive to a consensus in which paths to follow at any bifurcation. One heuristic would be to analyse a subset of all the possible paths, choose the k-gradual and feasible paths and vote the final solution. That would be Coherent Social Inclusion: the Less-discriminating Majority Winner Selection Rule (LDM-WSR).

\section{Future Work}\label{future-work}

The main goal of this paper is that $LDM-WSR$, namely the Less-discriminating Majority Winner Selection Rule ($LDM-WSR$) would be included in each Constitution, \emph{i.e.} each core of constitutive rules establishing the basis of every (Human) Society.

For the time being, Social Choice Optimization opens new opportunities for multidisciplinary research: from theoretical to applications in almost every field of Knowledge, going through all the possibilities in between as I would briefly introduce;starting though with the mention of recent work \cite{KELLY2020}\cite{DUDDY202025} that might seem hopeful for characterising the $LDM-WSR$ as Pareto.

\subsection{Policy-making Opmitization}

One could integrate the problem, constraints on Social Power and methods presented in a Linear Programming problem to maximize Social Power and minimize Social Discrimination. The author expects to follow in short-term this research path using Neuro-dynamic programming \cite{bertsekas1995neuro} from a Simulation-based Optimization perspective where a vehicle routing problem\cite{secomandi2000comparing} is extended with low-social power pedestrians and, (cross-walk or traffic light) norm allocations to solve this Coherent Social Inclusion problem, from a (toy) single-cross problem to a (real) world-wide problem.
Although promising results are presented in \cite{zhao2019event} and \cite{yang2020event}, the advantages of the proposal of this paper are that (norm) preferences are explicit and known for the problem to be solved contrary to the black-box approach of Neuro-dynamic programming for each function to be learned. However, a preference history (policy) could be found for each voting step with the mentioned approach.

\subsection{Abstraction and Aggregation in $\mathcal{L}^{p\rightarrow+\infty}$}

Defining a computer program, norm, constraint solving and so on is finding and constructing a function in the Lebesgue space ${\mathcal{L}^p}$. By defining the Social Discrimination function as a functional abstraction in Lebesgue space opens the research path of Computational Abstraction (definition, formalisation, characterization, implementation$\ldots$), maybe be as functions in hyper-connected $\mathcal{L}^{+\infty}$ Domains, involving all the Artificial Intelligence fields: Logics, Reasoning, Learning, Multi-agent Systems, Declarative Mechanism Design, and so on.

As a starting point, functions in different Domains (topics) define a new super-domain. Thus, I refer to super-domains created by two functions created with Lambda calculus. Then, recursively, using Lambda Calculus of two (or more) of these super-domains, and so on. I would name these two Lambda calculus of super-domains: Standardization and Integration.

\subsection{Characterization of Social Choice Optimization}

Starting in a top-down approach, I would mention than the most prominent theoretical research path would be the formalization of Social Choice Optimization properties. From checking what paradoxes apply and how much Condorcet Efficient would be. I followed a Hill climbing (citation following) approach starting from \cite{cond-eff2011} to \cite{incomplete-prefs2020} and the path has been at least interesting.

\subsection{Improving Coherent Social Inclusion}

\subsubsection{Learning Social Discrimination Profile}
The Social Discrimination Profile functions $SDP$ are discrimination estimations agreed upon previously over the Social Role of a Society. I have assumed than Discrimination may be analysed quantitatively. However, this is still an ongoing research path as \cite{low-power} and \cite{social-medicine} present.
This poses the next research question: Could Regulated Deep Learning \cite{agc:aicomm20-2} learn a discrimination function from empirical tests?

\subsubsection{Relaxing the number of traits and social profiles being discriminated}

In the One-stage Approval Voting algorithm for Coherent Social Inclusion, all the preferences of the Social Universe are taken into account. One research path would be characterizing and proposing solutions for the relaxed case where the number of traits, or social profiles, being discriminated is not an empty set.

\subsection{Increasing Uncertainty: Collaborative Knowledge Exploration}

Continuing with the Mountain Exploration metaphor, where I assumed all the kidnapped members of the Society are together in the same point of the Mountain and must remain together. Let me then assume a variation where the agents are scattered all over the Mountain, namely Discrimination Space. Then, one could start assuming that there exist Free Communication and the global (and common) social goal of everyone reaching at the highest peak, is maintained. In this case, the application domain is multidisciplinary.

\subsubsection{Assuming Agent Rationality and Will}

Following the extra definitions, we may find $\overrightarrow{K}_i$ the (Possible and Coherent) Knowledge function over a set of preferences $P_i$ and a set of Preference Discrimination functions ${PD_i}$ for agent $i$. We may define and implement the Agent Preferences function $K_i$ strategically following a Game Theory approach, dividing the Strategic Preference (and Preference Discrimination) Selection problem in possibly continuous steps, like continuous time.

\subsection{Implementing Coherent Social Inclusion in Declarative Mechanism Design.}

Following the work from \cite{agc:aicomm20-2} on Declarative Mechanism Design, where \I\ is improved, the author expects to start the Open Integration of Coherent Social Inclusion in the fore-mentioned Regulated Middleware.

Furthermore, the work presented previously in \cite{garcia2010normative} has been tested in the Electronic Institutions (\eis) Middleware presented in \cite{estevaphd} that runs over JADE \cite{jade}. Nonetheless, the author expects to add the language to a newly developed Middleware, as it is capable of give Operational Semantics, \emph{i.e} run, a Declarative version of \ei\ protocols as it was shown in \cite{garcia2010normative}. The main advantage of a Declarative version is that it might be provided to Software Agents as the rules-of-the-game in order to coordinate, \emph{e.g}, using Social DCOPs \cite{socialDCOP12}, AMODCOPs \cite{matsui2015leximin} or Game Theory. 

\subsection{Improving Game Theory}

From the complete implementation of Declarative Electronic Institutions found in \cite{garcia2010normative} we have just used one Activity, a Metamorphic Game, \emph{i.e.} a game whose rules may vary with time, action (or inaction) of agents, or other normative notions. To the best of my knowledge there are several types of Games not formalised yet in order to fully coordinate agents in a Declarative (Complete) Electronic Institution. Thus, there might be a need to formalise Hierarchical Concurrent Metamorphic Game Theory. However, following the Divide-and-Conquer methodology a first attempt of roadmap would be in the inverse order:

\[Metamorphic \Rightarrow Concurrent \Rightarrow Hierarchical\]

\[Metamorphic \Rightarrow Concurrent\ Metamorphic \]

\[Concurrent\ Metamorphic \Rightarrow\]
\[ Hierarchical\ Concurrent\ Metamorphic\]

Then, it would reasonable to study their interrelationships, thus creating Declarative Meta-Games, as the computable with the Declarative Electronic Institutions tested.

\subsection{Full-Hybrid Artificial Intelligence}
However, using $\I$ as an Programmable Event-based Middle-ware opens new paths of research as it uses Hybrid AI (\i.e. mixes Autonomous Agents and Multi-agents Systems, Machine Learning, and Symbolic Programming). Furthermore, one of the main applications of Hybrid AI is in its own a whole new AI subfield, namely, Artificial Teaching. 

\subsection{Artificial Teaching}

The whole concept of Artificial Teaching is recent, and not properly defined yet. There are some mentions in the literature that I will not cite in order to engage the reader to improve the previous lines and the proposed concepts luckily exposing her results on subsequent articles. For a sample, the reader may check ongoing research in \cite{teaching1} and \cite{teaching2}.

In my humble opinion, in the research path towards General AI there are several Problem-specific milestones to reach in every sub-field of AI; and mimicking Human Intelligence and Evolution, it may seem a natural step forward to add the teaching capability to artificial learners to decrease complexity. 

Please imagine a researcher (agent) being in a continuous "Deep... and deep... and deep... and very deep... Learning" process since the beginning of its existence. To the best of my knowledge, there are very few (human) researchers (honestly, almost none) that self-learned everything on his own, with no interaction with others who may have taught him something, even involuntarily, and this happens almost every day as a Spanish proverb well says. 

\subsection{Collaborative Knowledge Evolution}

Sumarizing my main goal as Collaborative Knowledge Evolution, it could be seen as the Open Standardization and Integration of Open Knowledge optimized from generation to generation thanks to (Human) Evolution.
With this, I want to emphasize the role of (Human and Artificial; Physical and Software) Teachers in Collaborative Learning and (Collaborative) Research, and thus in Collaborative Knowledge Evolution. Luckily, in a future we would be a step closer to General AI, by means of  Collaborative Optimisation, achieving thus a full integration and consensus of researchers (and their contributions, either Physical or Software), even they are not collaborating on purpose. And all these thanks to Regulated Middle-wares and Artificial Mediators.

\bibliography{aicomm.bib}
\bibliographystyle{ios1.bst}


\end{document}